# AGGA: A Dataset of Academic Guidelines for Generative AI and Large Language Models

Junfeng Jiao [1], Saleh Afroogh[2], Kevin Chen [3], David Atkinson [4], Amit Dhurandhar [5]

1. Urban Information Lab, The School of Architecture, The University of Texas at Austin, Austin, TX 78712, United States. jjiao@austin.utexas.edu
2. Urban Information Lab, The School of Architecture, The University of Texas at Austin, Austin, TX 78712, United States. Saleh.afroogh@utexas.edu
3. Department of economics, The University of Texas at Austin, Austin, TX 78712, USA. xc4646@utexas.edu
4. Allen Institute for AI (AI2), Seattle, USA davida@allenai.org
5. IBM Research Yorktown Heights, USA, adhuran@us.ibm.com

Corresponding author(s): Saleh Afroogh (Saleh.afroogh@utexas.edu)

## Abstract

This study introduces AGGA, a dataset comprising 80 academic guidelines for the use of Generative AIs (GAIs) and Large Language Models (LLMs) in academic settings, meticulously collected from official university websites. The dataset contains 188,674 words and serves as a valuable resource for natural language processing tasks commonly applied in requirements engineering, such as model synthesis, abstraction identification, and document structure assessment. Additionally, AGGA can be further annotated to function as a benchmark for various tasks, including ambiguity detection, requirements categorization, and the identification of equivalent requirements. Our methodologically rigorous approach ensured a thorough examination, with a selection of universities that represent a diverse range of global institutions, including top-ranked universities across six continents. The dataset captures perspectives from a variety of academic fields, including humanities, technology, and both public and private institutions, offering a broad spectrum of insights into the integration of GAIs and LLMs in academia.

## Background & Summary

Recent developments and deployments of artificial intelligence (AI) in academia, particularly generative AI (GAI) technologies like Large Language Models (LLMs), have sparked both excitement and concern. The increasing relevance and widespread adoption of GAI/LLMs in academic contexts have prompted diverse responses from universities worldwide. Some institutions, for instance, have resorted to traditional pen-and-paper exams to maintain academic integrity, while others have updated their plagiarism policies and AI-use guidelines [1]. Despite the growing adoption of GAI applications, a UNESCO global survey revealed that fewer than 10% of schools and universities have developed formal policies or guidance specific to the use of these technologies [2]. This gap highlights the urgent need for comprehensive frameworks that balance innovation with ethical considerations in the use of LLMs and other GAI tools within academia.[3][4][5]

GAI technologies, such as LLMs, utilize patterns learned from vast datasets to generate new content, including text, images, and music. LLMs are prominent examples of GAI. They are advanced neural network architectures designed to understand and predict probability distributions within linguistic sequences, fundamentally altering the way we interact with and produce written content.

The field of Natural Language Processing (NLP) has seen significant advancements with the emergence of GAI and LLMs. LLMs, characterized by their vast number of parameters—ranging from tens of millions to billions—are trained on large volumes of diverse data, enabling them to capture intricate linguistic patterns and generate coherent, contextually appropriate text [6]. Functioning in an autoregressive manner, LLMs predict the probability distribution of a token based on preceding tokens, leveraging the chain rule of probability and conditional probabilities. The generative nature of these models allows for the production of human-like text, showcasing sophisticated neural architectures that enhance contextual understanding, semantic coherence, and adaptability. These advancements have profound implications across various domains, particularly in academia, where their impact on teaching, research, and knowledge dissemination is increasingly recognized [6], [7].

Guidelines for the usage of GAIs and LLMs in academia are typically expressed using the most flexible communication code, which is natural language (NL). However, applying natural language processing (NLP) techniques in guideline engineering enables scholars to address a variety of tasks, including model synthesis, classification of requirements into functional/non-functional categories, ambiguity detection, structure assessment, and information extraction. Moreover, the lack of an inclusive text-based dataset forces most institutions to rely on proprietary or specific documents as benchmarks, hindering the replication of experiments and the generalization of results.

This study introduces AGGA (Academic Guidelines for Generative AIs), a dataset comprising 80 academic guidelines from various continents and countries, meticulously collected from their official websites. This dataset is designed to facilitate the replication of NLP experiments and the generalization of results. The documents within the dataset span multiple domains, offer varying degrees of abstraction, and include materials ranging from practical standards and public documents to university projects.

To validate the technical quality and inclusiveness of the AGGA dataset, we conducted two qualitative analyses as part of its comprehensive description. These analyses provide a detailed overview of the dataset's ability to support NLP-based studies of academic guidelines. The first analysis highlights the dataset's broad coverage across six continents, while the second, focusing on text-based analysis and visualization, examines different types of guidelines. Furthermore, we provide recommendations for effectively utilizing the dataset and suggest future expansions, including the addition of new official guidelines, which will be made available on GitHub for wider access.

**Figure 1.** Schematic Overview of the AGGA Study Design: Data Collection, Structuring, Analysis, and Utilization

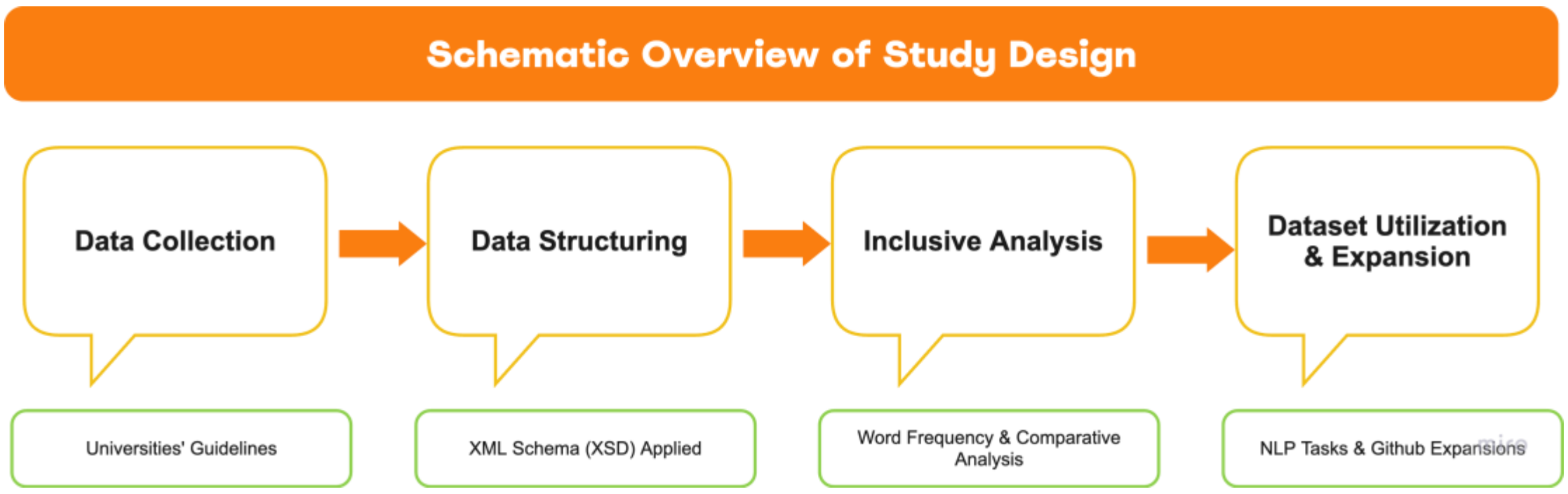

Figure (1) provides a clear and aesthetically pleasing representation of the key stages in your research:

1. Data Collection: Gathering guidelines from 80 universities across different continents.
2. Data Structuring: Applying an XML Schema (XSD) to standardize the documents.
3. Inclusive Analysis: Conducting word frequency analysis and assessing document structure.
4. Dataset Utilization & Expansion: Using the dataset for NLP tasks and planning to extend it by adding new guidelines on GitHub.

## Methods

In this study, we conducted a comprehensive data collection of academic guidelines for the use of Generative AIs (GAIs) and Large Language Models (LLMs) in academic settings. Our data collection phase focused on gathering official guidelines from 80 universities worldwide, ensuring a representative sample across different geographical regions and operational levels. The selection criteria for universities included top-tier status, geographical diversity, cultural representation, and the presence of official guidelines specifically addressing the use of GAIs, with an emphasis on LLMs. Universities lacking such guidelines were excluded, and replacements were selected based on similar criteria to maintain consistency and reliability in our analysis. This meticulous selection process was critical in ensuring the credibility of our findings (See, Table 1)

**Table 1.** Analytics Workflow of the analysis of GAI/LLM guidelines

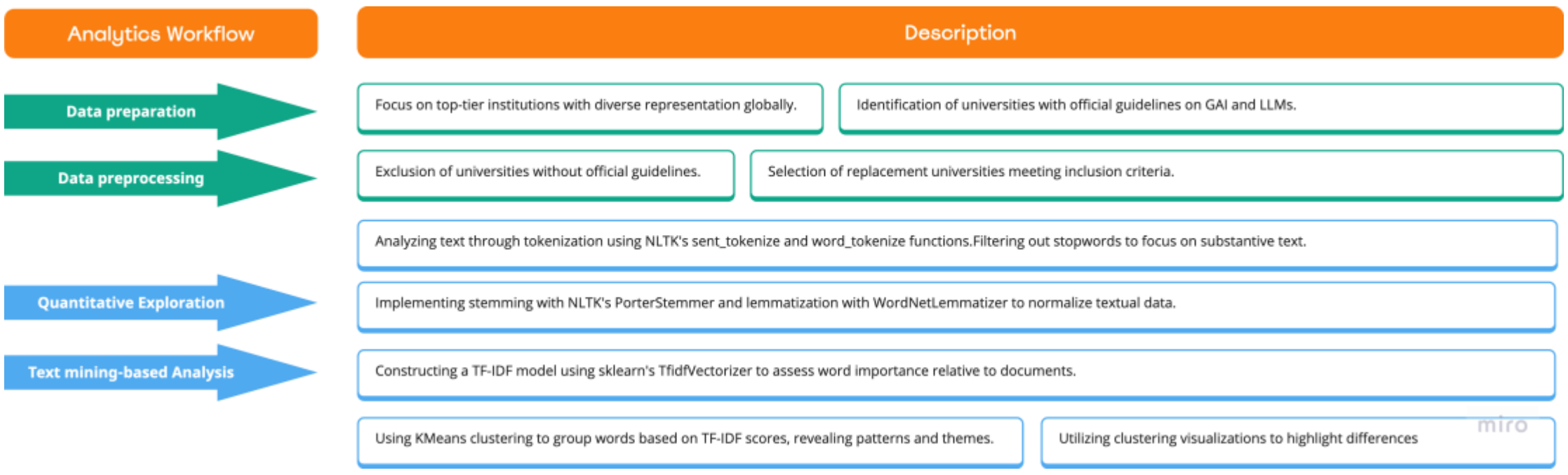

### Categorization Criteria

To gain an understanding of the current landscape of recommendations, two primary categorization criteria were used: geographical location and operational levels. Geographical location exploring the influence of regional factors on guideline development and application. We recognize the impact of cultural backgrounds and regional factors on the formulation and implementation of guidelines. Additionally, we referenced departmental guidelines from various regions, such as those from European and Asian countries, considering the regional specific use of AI tools like ChatGPT.

Operational levels, particularly at the university and departmental levels, were also considered when understanding the practical aspects, challenges, and variations in guideline implementation. Supplementary Table 1, presents 80 official guidelines collected from universities in different countries across six continents, reflecting a wide range of cultural perspectives and approaches (see Supplementary Table 1).

### Text Mining and Computational Processing

To analyse the collected guidelines, we employed a text mining-based approach. Utilizing the Python programming language and its myriad of packages, including the many tools from Natural Language Toolkit (nltk) for text-processing. The analysis began with tokenization, where the text data from the 80 guidelines was broken down into sentences and subsequently into individual words. By using sent_tokenize and word_tokenize functions from the nltk package, we can contextually simplify the text at a granular level, making them individual words.

To refine the data further, we filtered out stopwords using nltk's stopwords list, thereby eliminating commonly used words that add little semantic value. An additional list of added_stopwords were manually entered to provide a more relevant result when doing frequency analysis. This process reduces noise, providing a substantive text for further analysis.

Following tokenization, we implemented stemming and lemmatization to normalize the textual data. Stemming, performed using nltk's PorterStemmer, reduced words to their root form, grouping different forms of the same word. Lemmatization, using WordNetLemmatizer, converted words into their dictionary form.

### Data Exclusion and Ethical Considerations

Universities that lacked official guidelines specifically addressing the usage of GAIs and LLMs were excluded from this study. These exclusion criteria were established prior to the study to ensure the methodological consistency and reliability of our findings. Replacement universities were carefully selected to maintain the representativeness of the sample.

This study did not involve human or animal subjects, so ethical approvals were not required. The data collected and analyzed were all publicly available from official university websites, ensuring that no sensitive or confidential information was utilized in this research.

### Computational Tools

The computational processing of text data was conducted using Python 3.8, with key libraries including nltk for natural language processing, matplotlib and networkx for their data visualization capabilities, with standard libraries such as pandas, numpy, itertools, re, docx2txt, that assists in reading and processing our dataset. The text mining and processing methods were performed on a high-performance computing environment.

## Data Records

The AGGA dataset, titled "AGGA: A Dataset of Academic Guidelines for Generative AIs," is available on the Harvard Dataverse repository. It is associated with the Persistent Identifier doi:10.7910/DVN/XZZHA5 and was published on June 4, 2024. The dataset is available in its fourth version, contributed by scholars from the University of Texas at Austin, the Allen Institute for AI, and IBM Research. The dataset is licensed under the Creative Commons CC0 1.0 Universal Public Domain Dedication (CC0 1.0), allowing for unrestricted use and distribution, provided proper credit is given.

### Overview of Data Files

The AGGA dataset is presented in three distinct file formats: MS Word, PDF, and MS Excel, each serving different analytical purposes. The MS Word file, titled "AGGA Dataset of Academic Guidelines for Generative AIs.docx," is a 14.6 MB document published on June 6, 2024. It contains three sections: a table listing the 80 academic guidelines, academic citations for each guideline, and the complete text of all 80 guidelines, organized with a detailed table of contents. Similarly, the PDF file, "AGGA Dataset of Academic Guidelines for Generative AIs.pdf," mirrors the structure of the Word document, offering a 5.3 MB file format suitable for easy access and reference. This file was published on June 17, 2024. Finally, the Excel file, "AGGA.xlsx," is an 82.2 KB spreadsheet published on June 4, 2024, providing a structured dataset that includes details such as Guidelines ID, Continent, Country, University, Name of Document/Website, and Number of Pages. Each file format caters to different user needs, from comprehensive textual analysis to structured data extraction.

### File Structure and Format Details

The AGGA dataset files are meticulously organized to ensure clarity and ease of use for researchers and practitioners. The MS Word and PDF files provide a comprehensive textual representation of the academic guidelines, with each document sectioned for clarity. These files include a detailed table of contents, making navigation through the document straightforward. The content is organized into three primary sections: a table listing the academic guidelines, citations for each guideline, and the full text of the guidelines. This organization supports both in-depth qualitative analysis and quick reference.

The Excel file offers a more granular view of the dataset, categorizing the guidelines based on geographical and institutional parameters. This file is particularly useful for quantitative analysis and cross-referencing with other datasets. The structure of the Excel file allows for easy sorting and filtering, enabling users to conduct specific analyses based on their research needs. The combination of these formats ensures that the AGGA dataset is versatile and accessible for a wide range of academic and research purposes.

### Data Citation

To ensure proper attribution and to adhere to academic standards, each data file associated with the AGGA dataset should be cited as follows: "Jiao, Junfeng; Afroogh, Saleh; Chen, Kevin; Atkinson, David; Dhurandhar, Amit, 2024, 'AGGA: A Dataset of Academic Guidelines for Generative AIs', https://doi.org/10.7910/DVN/XZZHA5, Harvard Dataverse, V4." [8]

## Technical Validation

To validate the technical quality and inclusiveness of the AGGA dataset, we conducted two qualitative analyses as part of the comprehensive description phase of the dataset. These analyses aim to present a descriptive overview of the dataset's capacity to support NLP-based analyses of academic guidelines. The first analysis focuses on the inclusiveness of the dataset, demonstrating its comprehensive coverage across six continents. The second analysis involves a text-based analysis and visualization of the guidelines based on different types of guidelines, which will be detailed later in this section.

### Text-Based Analysis and Visualization Across Six Continents

The AGGA dataset includes academic guidelines from six continents: Africa, Asia, Europe, North America, South America, and Oceania. The textual data for these guidelines was extracted from the main AGGA dataset document, which is available in DOCX format. The preprocessing steps were executed using Python, with all necessary packages installed in Jupyter Notebook.

The initial step involved converting the DOCX files to plain text using the docx2txt library. This conversion ensured that the text data was in a standardized and easily readable format for subsequent processing. The extracted text content was then subjected to several preprocessing steps to prepare it for analysis.

### Text Processing

The text data underwent multiple preprocessing stages, including reading and lowercasing, symbol and number removal, tokenization, stopword elimination, stemming, and lemmatization. Initially, the text content was read from the converted TXT files and transformed to lowercase to maintain consistency. Non-alphanumeric characters and numbers were removed to clean the data further, eliminating elements that do not contribute to the text's semantic meaning.

The cleaned text was then tokenized into sentences and words. Sentence tokenization divides the text into individual sentences, while word tokenization splits these sentences into individual words or phrases. Stopwords, which are commonly used words that add little significance to the analysis (e.g., "and," "the," "is"), were removed to focus on more analytically relevant words. Additional keywords were also filtered out to ensure that the most relevant terms were highlighted, this list includes the words: 'use', 'used', 'using', 'may', 'work', 'also', 'text', 'ways', 'uio', ‘tool’, ‘university’, ‘provide’, ‘relevant’, ‘different’ ‘genai’, ’school’, ‘allowed’.

Stemming was applied using the PorterStemmer from nltk.stem , which reduced words to their root forms, grouping variations of the same word under a single root. Lemmatization was then performed using the WordNetLemmatizer from nltk.stem, simplifying words to their most basic form, cleaning and normalizing the dataset, with the processed text stored in separate files for further analysis.

### Frequency Analysis and Keyword Extraction

Following preprocessing, we identified the twenty most popular keywords for each continent. This step involved tallying the occurrences of each word after stopwords were removed and selecting the top words with the highest frequencies. These keywords were then visualized using frequency charts using the matplotlib library, which graphically displayed the most frequently occurring words for each continent. It is also important to note that words from the added stopwords list were removed purposefully due to it being less relevant when highlighted in visualizations, not due to their frequency counts being excluded from the top 20 words displayed.

**Figure 2**: The Frequency of Key Concepts in six continents

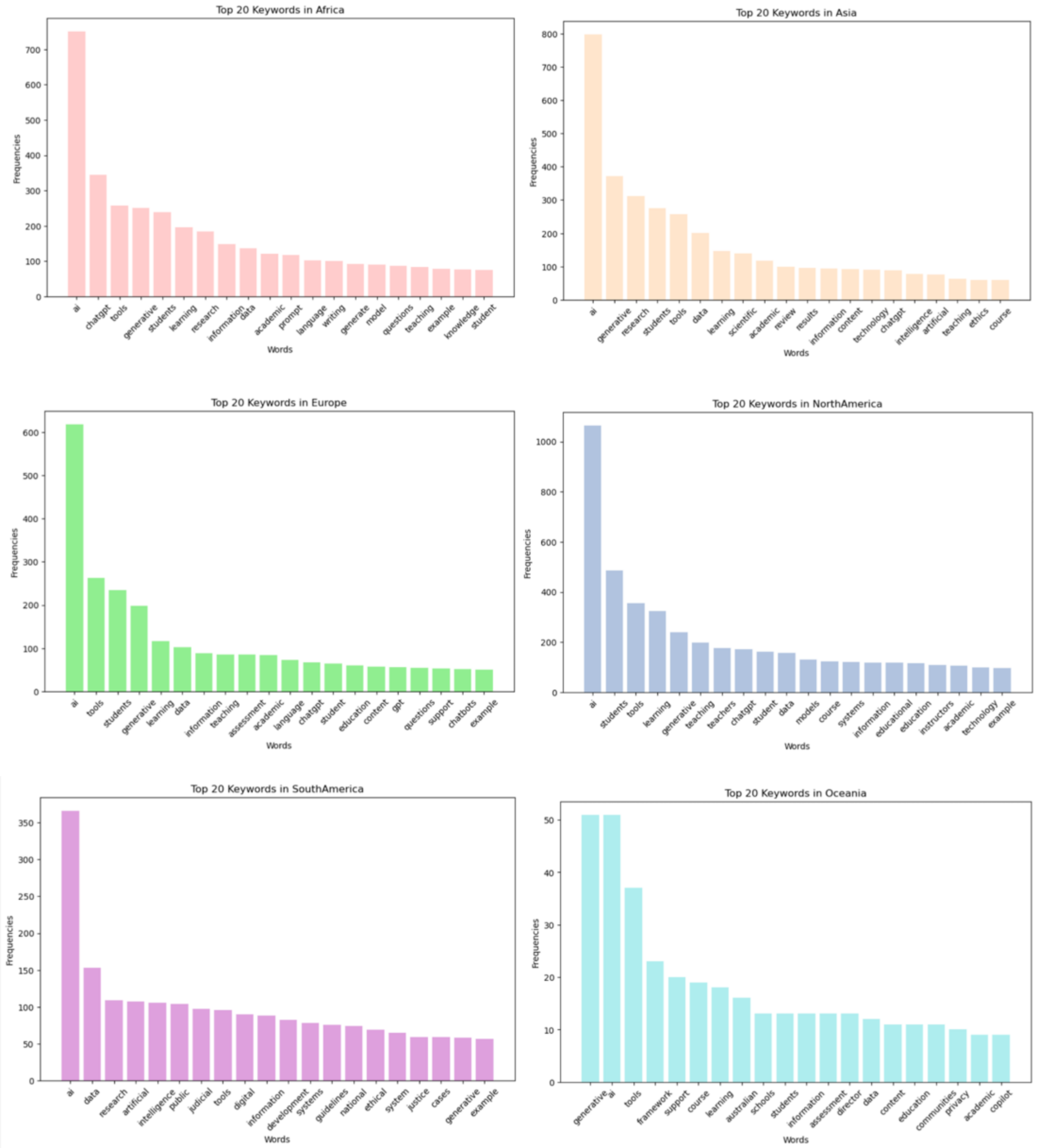

## Network Analysis

To further explore the relationships between the main themes (continents) and their associated keywords, a network graph was constructed using the NetworkX library. The process involved initializing a network graph, adding nodes and edges, assigning node attributes, and visualizing the graph.

Major nodes representing continents were added to the graph, with each node corresponding to a continent, and edges were created to connect these nodes to their top 20 associated keywords, along with minor nodes representing the top ten keywords for each continent. Edges were created to connect each continent to its associated keywords. keyword frequencies are calculated using Counter, which helped in determining the size of each minor node (keyword). These minor nodes were scaled based on frequency, with more frequent keywords given larger node sizes.

Attributes were assigned to both major and minor nodes to better predict their appearance and behaviour in the graph. Major nodes (continents) were assigned a larger base size of 1500 and given a red border to distinguish them from the minor nodes. Each continent node was also labelled with a “group” attribute, corresponding to the continent it represented. Minor nodes (keywords) were similarly assigned a “group” attribute matching the continent they were linked to, and their size was scaled based on frequency.

The nodes were positioned using the nx.spring_layout function from NetworkX. This layout function simulates physical forces, treating nodes as objects connected by springs. The k=0.5 and scale=2 parameters were used to better control the spacing between nodes. The spring layout was chosen because it more cosnsitently  to generate graphs where nodes with many connections are pulled closer together, making it easier to differentiate  relationships between continents and their keywords. With each node color-coded using that predefined color dictionary, assigning a specific color to each continent.

Edges were drawn to represent connections between each continent and its associated keywords. These edges were colored in gray and had an alpha value of 0.5 to make themtransparent, the use of light, neutral-colored edges makes sure focus remained on the nodes and their relationships, rather than the connections themselves. Red edge borders were used for the major nodes (continents), distinguish them from the minor nodes. Labels were added to both major and minor nodes using nx.draw_networkx_labels, and network graph is displayed using Matplotlib.

**Figure 3**: The Network Graph of Key codes and keywords in six continents

Network Graph of Major Codes and Keywords

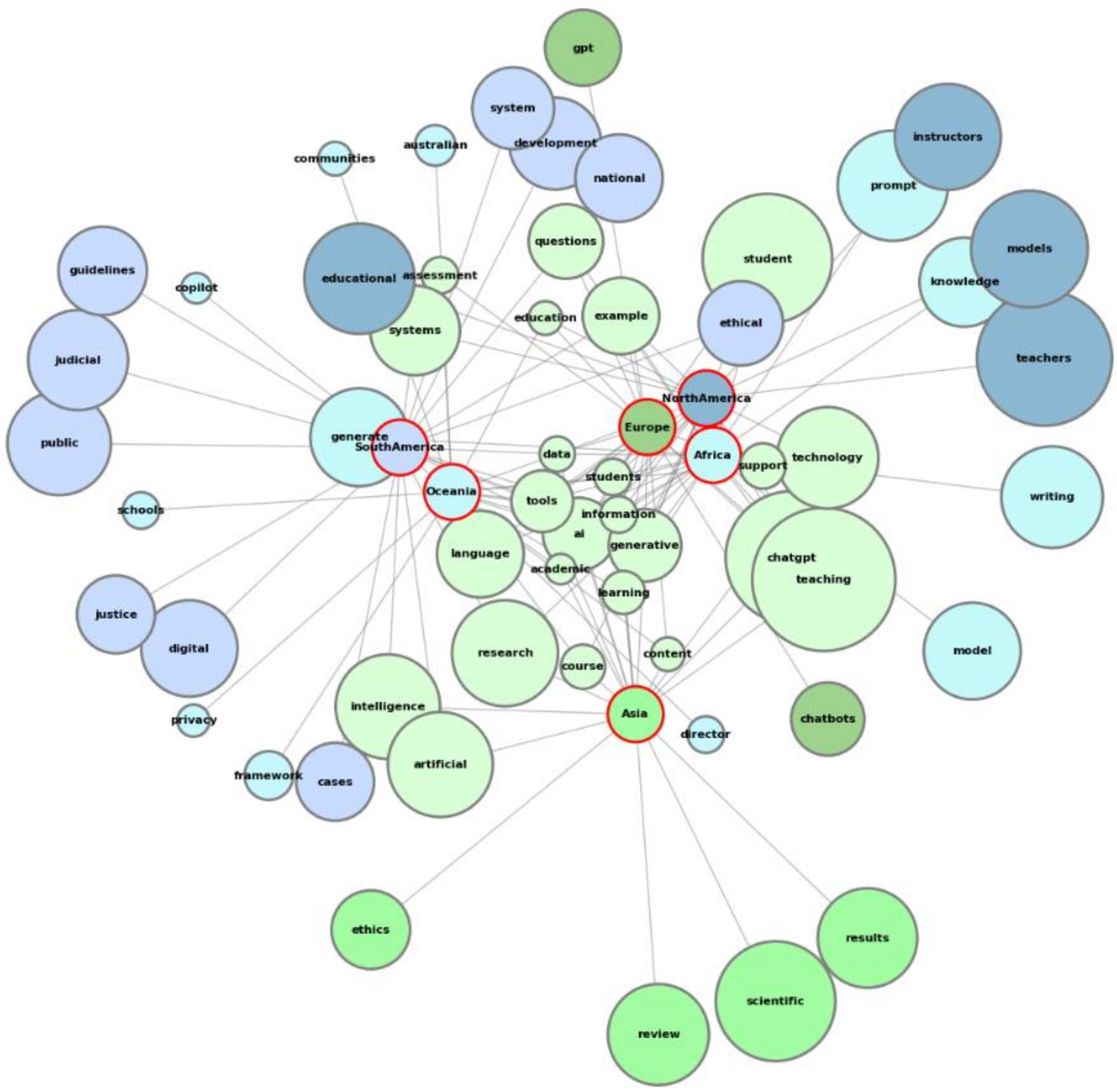

## Validation of Dataset Inclusiveness and Global Representation

The network graph effectively illustrates the connections between continents and their most relevant keywords, highlighting both distinct and shared themes across different regions. For example, terms such as "questions," "technology," "chatgpt," "teaching," "student," "research," "intelligence," "ai," and "artificial" are shared across multiple continents, reflecting common global themes in the use of AI in academia.

In Africa, the analysis reveals a strong emphasis on incorporating AI technologies into education, with keywords such as "chatgpt," "ai," and "tools" appearing frequently. This suggests a focus on using AI to enhance educational opportunities and academic research. In Asia, the analysis highlights a commitment to scientific exploration and research in AI, with "generative AI" and "research" being prominent keywords. Europe shows a wide spectrum of AI applications, with a particular focus on educational aspects, as indicated by keywords like "tools," "students," and "learning." North America is at the forefront of AI research and technology integration in education, with keywords such as "ai," "technology," and "systems" dominating the analysis. South America's analysis reflects a focus on AI applications in public domains and legal systems, with keywords like "public," "research," and "justice" standing out. In Oceania, the analysis highlights interest in AI technology and educational support, with terms like "generative AI," "frameworks," and "learning" being particularly relevant.

Overall, this text-based analysis and network visualization provide a robust validation of the dataset's inclusiveness and its potential to support comprehensive NLP analyses of academic guidelines on a global scale. The diverse themes and regional focuses identified in this analysis demonstrate the dataset's capacity to offer valuable insights into the integration of AI technologies in academic settings across different continents.

## Usage Notes

The AGGA dataset is designed to be a versatile resource for researchers and practitioners interested in exploring the academic guidelines for the use of Generative AIs (GAIs) and Large Language Models (LLMs) in academia. To facilitate the reuse of this dataset, we provide several suggestions and tips that can assist in maximizing its utility.

Researchers aiming to analyze the dataset can utilize a variety of software packages and tools. For text-based analyses, Python is highly recommended due to its comprehensive libraries, such as nltk for natural language processing tasks, sklearn for machine learning applications, and matplotlib or seaborn for data visualization. The dataset is compatible with standard text processing pipelines, and the provided DOCX, PDF, and Excel formats ensure that the data can be easily imported into most text analysis or spreadsheet software. We recommend several downstream processing steps for those interested in performing deeper analyses. For instance, researchers can implement normalization techniques to standardize the text data further, making it suitable for comparison across different guidelines. Additionally, topic modeling or sentiment analysis can be performed to uncover underlying themes or sentiments within the guidelines.

### Privacy and Safety Considerations

The AGGA dataset consists of publicly available academic guidelines collected from official university websites. Therefore, no privacy or safety controls are associated with public data access. Researchers are encouraged to use the data responsibly, adhering to the ethical guidelines relevant to their research field. Given the public nature of the data, there are no restrictions on access, and the dataset is freely available under the Creative Commons CC0 1.0 Universal Public Domain Dedication.

## Code Availability

The processing and analysis of the AGGA dataset were conducted using custom Python scripts, primarily within a Jupyter Notebook environment. The code used to convert DOCX files to plain text, preprocess the text data, and perform the text mining analyses is available upon request. Code is available in the GitHub account of corresponding author: https://github.com/SalehAfroogh/AGGA_Notebook.

The custom scripts were developed using Python 3.8 and relied on several key libraries, including nltk (version 3.5) for natural language processing, sklearn (version 0.24) for machine learning tasks, and matplotlib (version 3.3) for data visualization. Specific variables and parameters used in generating, testing, and processing the dataset, such as stopword lists, stemming rules, and lemmatization techniques, are documented within the code comments for clarity.

The code is intended to be reusable and adaptable for other researchers who may wish to replicate or extend the analysis. To ensure transparency and reproducibility, any updates or modifications to the code will be shared through the corresponding author's GitHub repository, with appropriate version control.

# Acknowledgements

This research was funded by the National Science Foundation under grant number 2125858. The authors would like to express their gratitude for the foundation's support, which made this study possible. Furthermore, in accordance with MLA, we would like to thank OpenAI for its assistance in editing.

# Author contributions

# Competing interests

The authors declare that the research was conducted in the absence of any commercial or financial relationships that could be construed as a potential conflict of interest.

**Supplementary Table 1**: University-level guidelines for Usage of LLMs and GAIs

| | Continent | Country | University | Name of document/website |
|---|---|---|---|---|
| **1** | **Africa** | South Africa | University of Cape Town | Artificial Intelligence for Teaching & Learning [9] |
| **2** | | Egypt | Cairo University | FCAI Policy and Guidelines for use of Generative AI in Postgraduate Studies and Research[10] |
| **3** | | South Africa | University of Witwatersrand | ChatGPT & other AI tools for Learning and Teaching[11] |
| **4** | | South Africa | University of Pretoria | Leveraging Generative Artificial Intelligence for Teaching and Learning Enhancement[12] |
| **5** | | South Africa | University of Johannesburg | UJ Practice Notes: Generative Artificial Intelligence in Teaching, Learning and Research[13] |
| **6** | | South Africa | Stellenbosch University | Stellenbosch University Academic Integrity: Responsible Use of AI tools[14] |
| **7** | | South Africa | University of the Free State | Stepping up with ChatGPT - AI-assisted Technology in Education[15] |
| **8** | | Egypt | The American University in Cairo | AUC's Statement on the Use of Artificial Intelligence Tools[16] |
| **9** | | South Africa | North-West University | Implications of AI for teaching and learning in higher education & Guidelines for the Utilization of AI in Teaching and Learning at NWU[17] |
| **10** | | South Africa | African Observatory on Responsible Artificial Intelligence | Generative AI guidelines at South African universities[18] |
| **11** | **North America** | USA | Harvard | Guidelines for Using ChatGPT and other Generative AI tools at Harvard[19] |
| **12** | | USA | Stanford | Generative AI Policy Guidance[20] |
| **13** | | USA | MIT | Getting Started with AI-Enhanced Teaching: A Practical Guide for Instructors[21] |
| **14** | | USA | Princeton | Generative AI Guidance[22] |
| **15** | | USA | University of Chicago | Guidance for Syllabus Staatements on the Use of AI Tools[23] |
| 16 | | USA | Columbia | Considerations for AI Tools in the Classroom[24] |
| 17 | | USA | California Institute of Technology | Guidance on the Use of Generative AI and Large Language Model Tools[25] |
| 18 | | USA | University of California, Berkeley | Appropriate use of ChatGPT and Similar AI Tools[26] |
| 19 | | USA | Yale | Guidelines for the Use of Generative AI Tools[27] |
| 20 | | USA | University of Pennsylvania | Statement on Guidance for the University of Pennsylvania Community on Use of Generative Artificial Intelligence[28] |
| 21 | | USA | University of California, Los Angeles | Guidance for the use of generative AI[29] |
| 22 | | USA | Cornell University | Cornell Gudelines for artificial intelligence[30] |
| 23 | | Canada | University of Toronto | Generative Artificial Intelligence in the classroom[31] |
| 24 | | Canada | University of British Columbia | Generative AI – Academic Integrity at UBC[32] |
| 25 | | Canada | McGill University | Principles on Generative AI in Teaching and Learning at McGill[33] |
| 26 | | Canada | University of Alberta | AI-Squared – Artificial Intelligence and Academic Integrity[34] |
| 27 | | Canada | University of Waterloo | Artificial Intelligence and ChatGPT – Academic Integrity[35] |
| 28 | | Canada | University of Montreal | Montreal Declaration on Responsible AI[36] |
| 29 | | Canada | McMaster University | Provisional Guidelines on the Use of Generative AI in Teaching and Learning[37] |
| 30 | | USA | U.S. Department of Education | Artificial Intelligence and the Future of Teaching and Learning[38] |
| 31 | **South America** | Colombia | Universidad del Rosario | Guidelines for the Use of Artificial Intelligence in University Courses[39] |
| 32 | | Argentina | University of Buenos Aires | Guidelines for the use of ChatGPT and text generative AI in Justice[40] |
| 33 | | Colombia | Universidad de Los Andes | Guidelines for the use of artificial intelligence in university contexts[41] |
| 34 | | Colombia | Pontifical Javeriana University | Editorial Policy, Publication Ethics and Malpractice Statement[42] |
| 35 | | Argentina | Universidad de san andres | Readiness of the judicial sector for artificial intelligence in Latin America[43] |
| 36 | | Peru | Government and Digital Transformation Secretariat | National Artificial Intelligence Strategy[44] |
| 37 | | Chile | Ministry of Science, Technology, Knowledge and Innovation | Guidelines for the use of artificial intelligence tools in the public sector[45] |
| 38 | | Chile | Pontificia universidad católica de chile | ChatGPT: How to use it in classes?[46] |
| 39 | **Asia** | China | Tsinghua University | International AI Cooperation and Governance Forum 2022[47] |
| 40 | | Singapore | National University of Singapore | Responsible Use of AI – Guidance from a Singapore Regulatory Perspective[48] |
| 41 | | Japan | Nagoya University | Regarding the Use of Generative AI[49] |
| 42 | | Japan | University of Tokyo | Guidelines for Instructors Regarding AI in University Education at Tokyo University of Foreign Studies[50] |
| 43 | | Hong Kong SAR | University of Hong Kong | Use of Artificial Intelligence Tools in Teaching, Learning and Assessments: A Guide for Students[51] |
| 44 | | South Korea | Seoul National University (SNU) | Seoul National University AI Policy Initative[52] |
| 45 | | Japan | Kyushu University | Note on the Use of Generative AI in Education at Kyushu University – For Teachers -[53] |
| 46 | | Japan | Weseda University | About the Use of Generative Arificial Intelligence (ChatGPT, etc.)[54] |
| 47 | | South Korea | Ulsan National Institute of Science and Technology | A Guide to the Use of Generative AI[55] |
| 48 | | Singapore | Signapore Management University | SUM Framework for the use of Generative AI Tools[56] |
| 49 | | Singapore | Singapore University of Technology and Design | Artificial Intelligence in Education[57] |
| 50 | | Singapore | Singapore Institute of Technology | Generative AI at the Singapore Institute of Technology[58] |
| 51 | | Singapore | Nayang Technological University | NUT Position on the Use of Generative Artificial Intelligence in Research[59] |
| **52** | | Taiwan | National Tsing Hua University | Guidelines for Collaboration, Co-learning, and Cultivation of Artificial Intelligence Competencies in University Education[60] |
| **53** | | Taiwan | National Taiwan University | Guidance for Use of Generative AI Tools for Teaching and Learning[60] |
| **54** | | Hong Kong SAR | The Chinese University of Hong Kong | Use of Artificial Intelligence Tools in Teaching, Learning and Assessments A Guide for Students[61] |
| **55** | | Thailand | Chulalongkorn University | Chulalongkorn University Principles and Guidelines for using AI Tools[62] |
| **56** | | Malaysia | Universiti Malaya | ChatGPT General Usage[63] |
| **57** | | Malaysia | Universiti Putra Malaysia | Guide for ChatGPT usage in Teaching and Learning[64] |
| **58** | | China | The Supervision Department of the Ministry of Science and Technology | Guidelines for Responsible Research Conduct (2023)[65] |
| **59** | **Australia** | Australia | The Department of Education | The Australian Framework for Generative Artificial Intelligence (AI) in Schools[66] |
| **60** | | New Zealand | The University of Auckland | Advice for students on using generative artificial intelligence in coursework[67] |
| **61** | **Europe** | UK | University of Oxford | Use of generative AI tools to support learning[68] |
| **62** | | UK | University of Cambridge | Artificial intelligence and teaching, learning, and assessment[69] |
| **63** | | UK | Imperial College London | Generative AI Guidance[70] |
| **64** | | UK | London School of Economics and Political Science | School Statement on Generative Artificial Intelligence and Education[71] |
| **65** | | UK | University College London | Using AI tools in assessment [72] |
| **66** | | UK | The University of Edinburgh | AI Guidance for Staff and Students[73] |
| **67** | | Netherlands | Erasmus University Rotterdam | Gnenerative AI in education[74] |
| **68** | | Belgium | KU Leuven | Responsible use of generative Artificial Intellgence[75] |
| **69** | | Switzerland | ETH Zurich | AI in education, resources for teaching faculty[76] |
| **70** | | Netherlands | University of Amsterdam | AI tools and your studies[77] |
| **71** | | Norway | University of Oslo | Guidelines to use artificial intelligence at UiO[78] |
| **72** | | Finland | University of Helsinki | Using AI to support learning[79] |
| **73** | | Italy | University of Padua | Research in Artificial intelligence[80] |
| **74** | | Sweden | Stockholm University | Guidelines on using AI-powered chatbots in education and research[81] |
| **75** | | Denmark | Technical University of Denmark | DTU opens up for the use of artificial intelligence in teaching[82] |
| **76** | | Netherlands | Delft University of Technology | AI chatbots in unsupervised assessment[83] |
| **77** | | Portugal | Universidade de Lisboa | Artificial Intelligence in education – Técnico presents resolution on the use of tools such as ChatGPT[84] |
| **78** | | Netherlands | University of Utrecht | Guidelines for the use of generative AI[85] |
| **79** | | Switzerland | University of Zurich | Guidelines for the Use of AI Tools[86] |
| **80** | | UK | Russell Group (24 UK research-intensive universities) | Russell Group principles on the use of generative AI tools in education[87] |

**Conflict of interest**: The authors declare that the research was conducted in the absence of any commercial or financial relationships that could be construed as a potential conflict of interest.

**Acknowledgements**This research is funded by the National Science Foundation (NSF) under grant number 2125858. The authors express their gratitude for the NSF's support, which made this study possible. Furthermore, in accordance with MLA (Modern Language Association) guidelines, we note the use AI-powered tools, such as OpenAI's applications, for assistance in editing and brainstorming.

**Institutional Review Board Statement:** Not applicable.
**Informed Consent Statement:** Not applicable.